\documentclass[conference]{IEEEtran}
\IEEEoverridecommandlockouts
\usepackage{cite}
\usepackage{amsmath,amssymb,amsfonts}
\usepackage{algorithmic}
\usepackage{graphicx}
\usepackage{textcomp}
\usepackage{xcolor}
\graphicspath{ {./img/} }
\def\BibTeX{{\rm B\kern-.05em{\sc i\kern-.025em b}\kern-.08em
    T\kern-.1667em\lower.7ex\hbox{E}\kern-.125emX}}
\begin{document}

\title{GAN pretraining for deep convolutional autoencoders applied to Software-based Fingerprint Presentation Attack Detection}

\author{\IEEEauthorblockN{1\textsuperscript{st} Tobias Rohrer}
	\IEEEauthorblockA{\textit{University of Applied Sciences Darmstadt} \\
		Darmstadt, Germany \\
		sttorohr@stud.h-da.de}
	\and
	\IEEEauthorblockN{2\textsuperscript{nd} Jascha Kolberg}
	\IEEEauthorblockA{\textit{University of Applied Sciences Darmstadt} \\
	Darmstadt, Germany \\
	jascha.kolberg@h-da.de}
}

\maketitle

\begin{abstract}
The need for reliable systems to determine fingerprint presentation attacks grows with the rising use of the fingerprint for authentication. This work presents a new approach to single class classification for software-based Fingerprint presentation attach detection. The described method utilizes a Wasserstein GAN to apply transfer learning to a deep convolutional autoencoder. By doing so, the autoencoder could be pretrained and finetuned on the LivDet2021 Dermalog sensor dataset with only 1122 bona fide training samples. Without making use of any presentation attack samples, the model could archive a average classification error rate of 16.79\%. The Wasserstein GAN implemented to pretrain the autoencoders weights can further be used to generate realistic looking artificial fingerprint patches. Extensive testing of different autoencoder architectures and hyperparameters let to coarse architectural guidelines as well as to multiple implementations which can be utilized for future work. 
\end{abstract}


\section{Introduction}\label{sec:1}
The usage of our fingerprint for authentication is getting more and more popular \cite{biometrics_use}. While this has several advantages like the ease of use, it also has a downside of being vulnerable to presentation attacks as shown in multiple studies \cite{vanderPutte2000, howtofake}. To mitigate this risk, different methods of performing presentation attack detection (PAD) have been developed. They can be subdivided into hardware and software based techniques. This document describes exclusively software based methods, which have the advantage of being more flexible as they do not require any additional sensor to be added to the fingerprint capture device.

Most of the PAD methods presented in the past used closed-set binary classification strategies where a model is trained using bona fide and presentation attack (PA) samples to be later able to discriminate between them. Another way of performing PAD is to use a single-class classification strategy, where a classifier is only learned using bona fide fingerprints. One-class-classifiers attempt to learn a representation of a live fingerprint to be able to detect everything but bona fide fingerprints. This has two major advantages:
\begin{enumerate}
	\item One-class classifiers do not overfit to a particular spoof material as they are solely trained on bona fide samples.
	\item For training a one-class classifier only bona fide samples are required. This saves a lot of time and effort when generating train data.
\end{enumerate}

While this sounds very promising in theory, learning one-class-classifiers like Generative Adversarial Networks (GANs) or Autoencoders (AEs) showed to be challenging in practice \cite{dcgan, gan_fingerprint}. This work presents a new method to one-class classification by using transfer learning to pretrain a convolutional AE using a Wasserstein GAN (WGAN). This approach combined with the usage of data augmentation enabled the AE to be pretrained and finetuned with very limited training data. The model is trained and validated using the LivDet2021 Dermalog sensor dataset. In summary, the main contributions of this work are:
 
\begin{enumerate}
	\item An implementation of a WGAN which can be used to generate synthetic fingerprints
	\item A proposal of a simple, yet effective method to pretrain a deep convolutional AE using a WGAN
	\item Architectural guidelines of building deep convolutional AEs, created by heavily experimenting with different architectures and hyperparameters.
	\item Experimental evaluation of classification capabilities of a one-class classification approach by solely using GANs
	\item Implementation of a input processing pipeline which applies data augmentation heavily which enabled the proposed concepts to work on a very limited training data set of only 1122 bonda fide fingerprints.
	\item TensorFlow implementations of several AE architectures including fully convolutional Autoencoders and variational convolutional Autoencoders, as well as implementations of a GAN and WGAN according to the DCGAN architectural guidelines \cite{dcgan}.
\end{enumerate}

The rest of the document is structured as follows: Section \ref{section:related} describes the current state-of-the-art approaches in the field of fingerprint PAD with an focus on one-class classification. The main concepts of the presented work are described in section \ref{sec:method}. Implementation details of these concepts are presented in section \ref{sec:implementation}. This is followed by section \ref{sec:experiments} where the results of the presented approach are presented. Furthermore experiments of using solely GANs for classification as well as training convolutional autoencoders from scratch are described in section \ref{sec:experiments}. A final conclusion as well as a listing of possible future work can be found in the final section.

\section{Related Work} \label{section:related}

\subsection{Fingerprint PAD as binary classification problem:}
A lot of research was done in the past where fingerprint PAD was trying to be solved as a binary classification problem. According to the results of most of the past LivDet challenges (2009, 2011, 2015, 2017), deep learning approaches using CNNs outperformed other approaches \cite{livdet1, livdet2, nogueira_old, cnn2017winner}. Nogueira et al. who won the LivDet-2015 challenge used transfer learning to retrain the CNN-VGG which was used to detect a PA based on the whole fingerprint image at once\cite{full_image}. In \cite{patch_based, korean} they also used a CNN but divided the fingerprint image into non overlapping smaller patches for training and classification. In \cite{minutae_based} they further optimized this procedure by extracting patches centered around minutiae points. This minutiae centered patch approach was used with transfer learning and archived state-of-the-art results on benchmark datasets of the LivDet competition. The best performing model of the LivDet 2019 challenge transformed each fingerprint image into three different image representations to combine local and global information of the fingerprint. These representations were transformed into a common feature space to train a linear SVM \cite{livdet_2019_winner}.

Besides the published results on fingerprint PAD, in the last semesters Biometric Systems course a CNN was implemented by applying transfer learning with the pretrained MobileNet V2 \cite{MobileNetV2}. The network was retrained within 160 epochs. By applying learning rate scheduling the CNN training started with a base learn rate of $10^{-5}$  within the first 60 epochs and gradually lower the learning rate during the rest of the training. Image augmentation (cropping, flipping, rotating and brightness adjustment) was not only applied during the training process but also during classification. 10 different random augmentations of the image to be classified were created. Every augmentation is classified individually and the average of all prediction scores forms the final score. The CNN uses the whole fingerprint image for classification and training at once. With this approach near Benchmark results on the LivDet 2015 CrossMatch sensor dataset could be archived. The network was submitted to the LivDet 2021 competition.
   
\subsection{One-class classification problem}
As described in section \ref{sec:1}, the theoretical advantages of an one-class classification approach are immense. As publications of one-class classification approaches in the Fingerprint PAD domain are comparably rare, this section also describes related work in the field of face PAD and related anomaly detection approaches.
\subsubsection{Non Deep learning approach}
An ensemble of One-class SVMs was trained on only bona fide samples using textural features like local binary patterns for fingerprint PAD \cite{one_class_svm}. They could validate their model with 15\% average classification error on the LivDet2011 Biometrika sensor data. In \cite{face_pad} they used an feature extractor using image distortion features \cite{iqm} to train a Gaussian Mixture Model based anomaly detector for face PAD.\\

\subsubsection{Autoencoders}
Another more common anomaly detection approach of using Autoencoders has also been applied to PAD lately. In \cite{ae_fingerprint} a convolutional Autoencoder has been successfully used for fingerprint PAD. They reported detection equal error rate of 2\%. It must be noted that the images used in their work were captured by a custom camera setup with multiple cameras and illuminations, which makes the classification for a model easier compared to the fingerprint images captured by the Dermalog sensor which were used in the presented work. Also the number of used training samples was almost 5 times higher than in the presented work.
In \cite{ae_face} several Autoencoders were pretrained using the CelebA dataset \cite{celebA} to be used as feature extractors for face PAD. A Multi layer perceptron (MLP) which uses the autoencoders latent vectors as inputs was used for classification. It must be noted, that this MLP was trained on bona fide as well as PA samples. Compared to the approach of the presented work, they still required some PA samples to train parts of their architecture.\\
 
\subsubsection{GANs}
GANs have been used lately for fingerprint PAD. In \cite{gan_fingerprint} three GANs similar to the DCGAN architecture \cite{dcgan} were used. They discarded the Generator after training and used solely the discriminator for classification. Their models were trained and validated by a custom data set which was created using a RaspiReader\cite{raspireader}, which makes it hard to compare their results. Further GAN based anomaly detection approaches are summarized in \cite{gan_anomaly_survey}. The AnoGan \cite{anogan} architecture utilizes a standard GAN architecture to be trained on only positive samples. By doing so the generator learns to generate only positive (non-anomalous) samples giving a latent encoding vector of fixed size. During classification they are trying to find an optimal latent encoding to generate an output using the generator, which is similar to the sample to be classified. As the generator was only trained on positive samples, the construction of positive samples should work better as of anomalies. The downside of their approach is that they have to find a good latent encoding for every sample to be classified which is computational expensive not only during training but also inference. The Efficient GAN-Based Anomaly Detection (EGBAD) \cite{efficient_gan_anomaly} on the other hand applies a BinGan \cite{bigan} architecture, by learning an encoder which is able to map inputs directly to the latent encodings directly during training. This extends the GAN architecture to have not only a Generator and Discriminator but also an Encoder. In \cite{ganomaly} a GAN and an AE were combined into one network architecture and later used for anomaly detection. Compared to \cite{ganomaly} and \cite{efficient_gan_anomaly} the approach presented in the work at hand has the advantage of utilizing the well established DCGAN architecture \cite{dcgan}. The DCGAN architecture was reported to work across different datasets, which is important as training of GANs architectures is a error prune task as it is highly sensitive to small changes in hyperparameter settings or of the data used \cite{dcgan, handsOnMachineLearning}.

\section{Proposed Method}\label{sec:method}
The main idea behind the proposed method is to utilize a GAN to pretrain weights for a deep convolutional autoencoder. The GANs discriminators weights are transferred to the autoencoders encoder as the discriminator already learned to transfer an input fingerprint image into lower dimensions for classification. In contrary, the GANs generators weights are transferred to the autoencoders decoder as the generator already learned to generate fingerprint images based on a latent input vector.

\subsection{Data Preprocessing Pipeline}\label{sec:preprocessing}
\begin{figure}[t]
	\centering
	\includegraphics[scale=0.15]{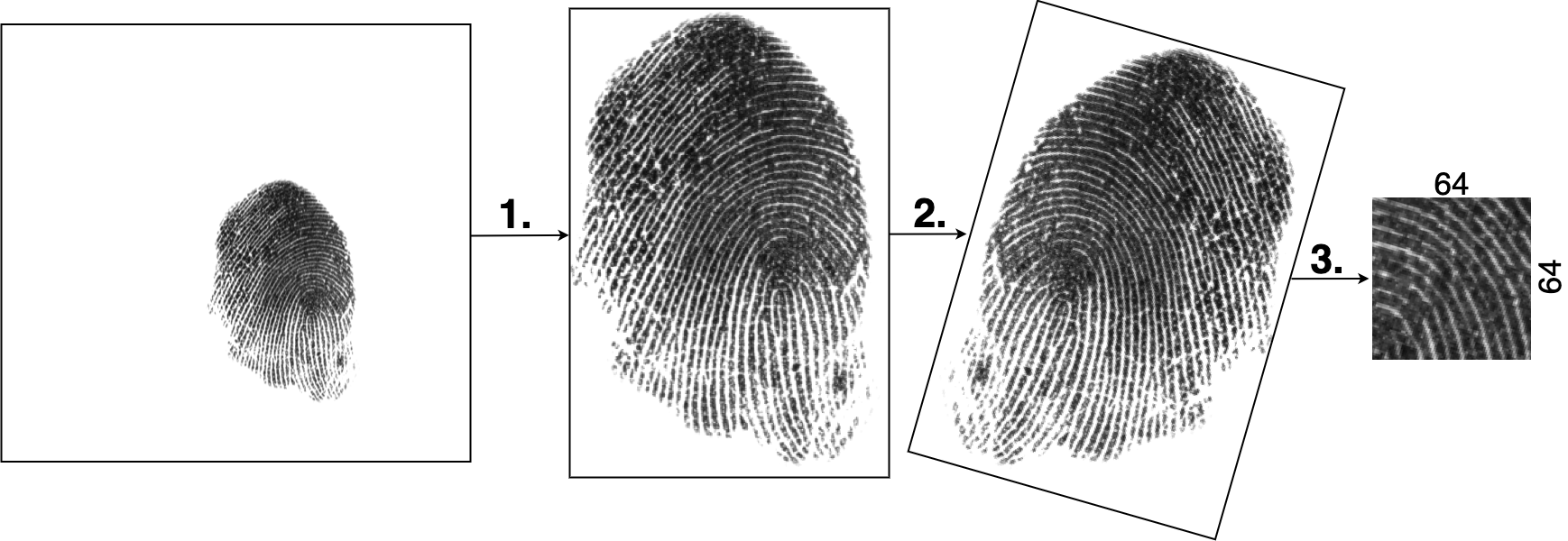}
	\caption{The three steps of the image preprocessing pipeline: 1.) ROI extraction 2.) Random data augmentation 3.) Random cropping}
	\label{fig:preprocessing}
\end{figure}
The pretraining of the weights using the GAN as well as the training of the autoencoder use the same simple yet effective image data preprocessing pipeline which is shown in Fig.~\ref{fig:preprocessing}. First, the region of interest (ROI) is extracted from the input image by simply iterating over the fingerprint image to remove surrounding empty rows and columns. Second, data augmentation is used to support training on the small amount of image data available. Each input image is being randomly vertical flipped, randomly rotated (between -20 and 20 degrees) as well randomly adjusted in the brightness by multiplying the image by a random factor (between 0.75 and 1.25). In the third and last step, a 64x64 patch of the fingerprint is randomly cropped from the augmented input image. It must be noted, that by randomly cropping from the input image, the diversity in training data can be further enhanced significantly. During validation and testing a slightly different pipeline is used. The first step remains the same to extract the ROI from the fingerprint image. Secondly the ROI is divided in multiple non overlapping 64x64 patches without any augmentation to be classified independently.

\subsection{Pretraining using a GAN}\label{sec:GAN_pretraining}
\begin{figure}[t]
	\centering
	\includegraphics[scale=0.28]{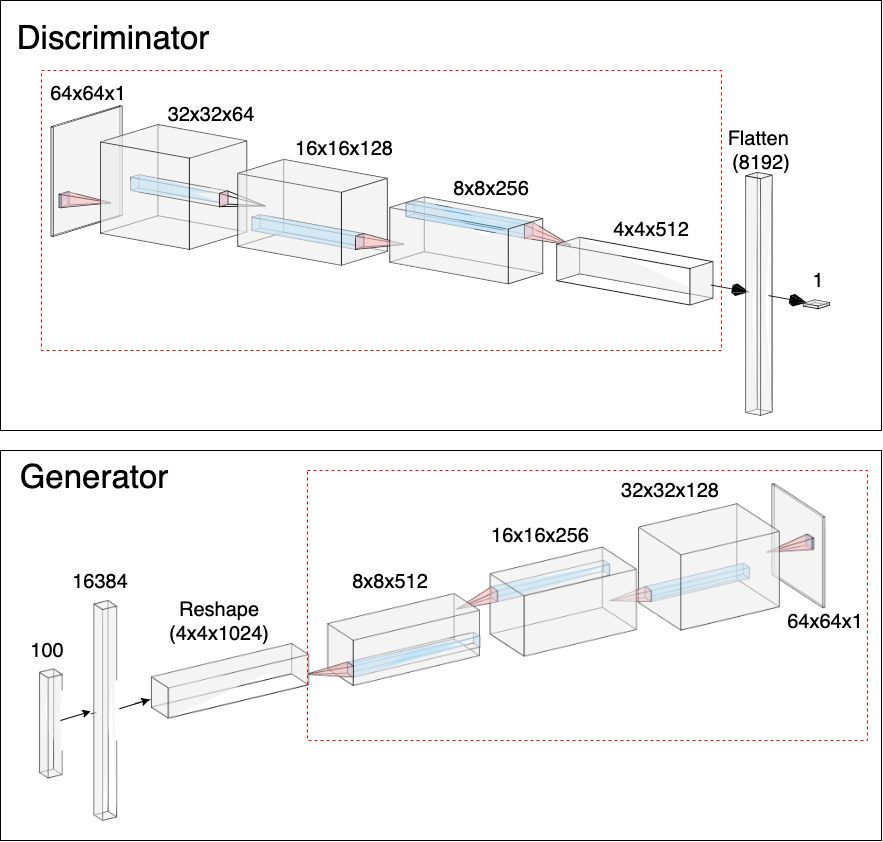}
	\caption{Generator and discriminator of the GAN. The red dotted boxes denote the parts of the GAN which are transferred to the AE.}
	\label{fig:gan}
\end{figure}

The concept of GANs were proposed 2014 in \cite{gan} as "a new framework for estimating generative models via an adversarial process, in which we simultaneously train two models". The two models the GAN consists of are the generator and the discriminator. The discriminator has the task to classify, if data was generated by the generator or if it came from the training data. The generators task is to try to fool the discriminator by generating new samples which are hard to differentiate from the training data. Hence the generator learns to create synthetic new samples without ever seeing a single training sample.

A GAN according to the DCGAN architecture \cite{dcgan} was implemented with the goal to apply transfer learning to a deep convolutional autoencoder with a similar architecture. The implemented DCGAN architecture is shown in Fig.~\ref{fig:gan}. All hyperparameters like activation functions, filter sizes, random weight initialization were selected according to the proposed methods in \cite{dcgan}. Still, the plain DCGAN implementation caused the GAN to run into mode collapse during training. Mode collapse happens when the generator is not able to learn a wide variety of outputs, but learns just a single or very few outputs to trick the discriminator. The usage of a Wasserstein loss function according to \cite{wassersteinGan} stabilized training and enabled the DCGAN to generate diverse synthetic fingerprints based on a latent input encoding. The hypothesis was, that a more stable GAN learns more divers and richer hence useful features, which can be later transferred to the AE. Following, the DCGAN with Wasserstein loss will be listed as WGAN and the plain DCGAN without Wasserstein loss simply as DCGAN.

\subsection{Convolutional Autoencoder}

An AE architecture typically contains an encoder and decoder. The encoder tries to learn efficient representations of the input data by transferring the data into a latent encoding. The decoder tries to reconstruct the input based on the latent encoding. The effectiveness of an AE is measured by a reconstruction error, which is a measure how well the input of the AE could be reconstructed by transferring it to a latent encoding and back. The AE is trained to minimize this reconstruction error \cite{handsOnMachineLearning}. When applying this concept to PAD the goal is to train an AE solely on bona fide fingerprint images, so it learns to encode and reconstruct them with a minimal reconstruction error. The hypothesis is, that PA samples which have not been used during training would have a higher reconstruction error and can be detected as PA.

\begin{figure}[t]
	\centering
	\includegraphics[scale=0.28]{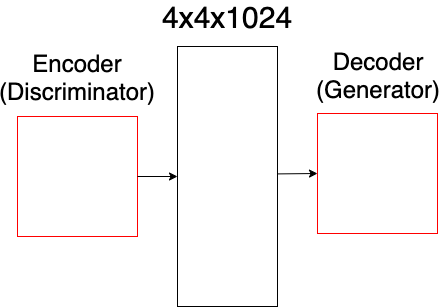}
	\caption{Autoencoder architecture. The red boxes denote the parts which are derived from the GAN architecture (Fig.~\ref{fig:gan}). They are connected by a convolutional layer with 1024 5x5 kernels, leading to a layer of size 1024x4x4.}
	\label{fig:ae}
\end{figure}

The architecture of the AE used in this work derived from the GAN architecture which can be seen in Fig.~\ref{fig:gan}. The red boxes in Fig.~\ref{fig:gan} denote the reusable parts of the GAN which are used similar in the AE architecture to apply transfer learning. The reusable part of the discriminator is used as encoder and the reusable part of the generator as decoder. A convolutional layer is added between both parts to archive the correct input size of the reusable generator part which is 4x4x1024. The final architecture of the AE is shown in Fig.~\ref{fig:ae}.

Training a deep convolutional AE with little training data is prone to overfitting or can lead to the problem that the AE does not learn any efficient latent encoding at all. This problem was solved by applying transfer learning by using a pretrained GAN with similar architecture as described in section \ref{sec:GAN_pretraining}, as well as heavy usage of image data augmentation. 

\section{Implementation details}\label{sec:implementation}
The proposed methods containing the preprocessing pipeline, the Wasserstein GAN, the convolutional AE as well as the experiments described in section \ref{sec:experiments} were implemented in Python using \emph{TensorFlow 2} \cite{tensorflow2015-whitepaper}. The preprocessing pipeline described in section \ref{sec:preprocessing} was implemented using the \emph{TensorFlow Dataset API}.

\subsection{Dataset}\label{sec:dataset}
The proposed models were trained using the LivDet2021 Dermalog Sensor dataset. The dataset contains only 1250 bona fide and 1500 PA fingerprints. As the official testing datasets for the LivDet2021 challenge are not available yet, 128 bona fide and 128 PA fingerprints were used for tracking the training process as well as testing the performance of the trained model. It must be noted that the WGAN and the AE were both trained only on bona fide fingerprints, meaning that only 1122 bona fide samples were used for training. 

\subsection{Pretraining weights using WGAN}
The WGAN was trained from scratch using the train data described in section \ref{sec:dataset}. Both, the Generator and the Discriminator were trained using the adam optimizer \cite{kingma2017adam} with a learning rate of 0.0002 and the \emph{beta\_1} parameter set to 0.5 and \emph{beta\_2} parameter set to 0.9. The adversarial training of the WGAN was configured in a way that for each update step of the generator there are three update steps for the discriminator, which further improves stability during training \cite{wgan_impl, wassersteinGan}. The WGAN was trained for 110 epochs with a mini batch size of 64. The WGAN had to be trained on the local development machine because of TensorFlow incompatibility issues. The training took around 8 hours to complete on a Intel Core i9  CPU with 8 cores.

\subsection{Transfer Learning Autoencoder}
\begin{figure}[t]
	\centering
	\includegraphics[scale=0.65]{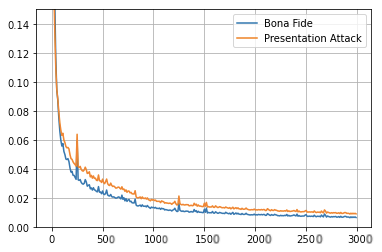}
	\caption{Reconstruction error on the validation data during the 3000 training epochs}
	\label{fig:training}
\end{figure}

First, the trained reusable parts of the Generator and Discriminator were transferred into the AE architecture as described in section \ref{sec:method}. The AE was then finetuned using the adam optimizer and a very low learning rate of $10^-5$ and the \emph{beta\_1} parameter set to 0.5 for 3000 epochs with at mini batch size of 32. The change of the reconstruction errors for bona fide and PAs during the training is shown in Fig.~\ref{fig:training}. The training was done on a Tesla M10 GPU and took around 25 hours to complete.

\subsection{Classification}\label{sec:classification}
For classification, the ROI is extracted from every test image and afterwards divided into non overlapping patches of size 64x64. A reconstruction error is being calculated for every patch individually by the AE. A final reconstruction error which is used for classification is being calculated by averaging all individual reconstruction errors. The classification threshold is defined by calculating the mean reconstruction error of all the training samples and adding their standard deviation. Samples with lower reconstruction errors than the defined threshold are classified as bona fide and others as PA. By defining the classification threshold like this, not a single PA sample is required during training.

\section{Results}\label{sec:experiments}

\subsection{Detection capabilities of the AE pretrained using a WGAN}\label{sec:detection}
\begin{figure}[t]
	\centering
	\includegraphics[scale=0.63]{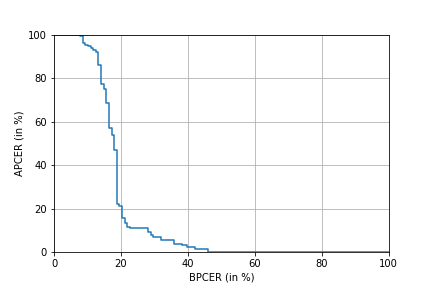}
	\caption{DET curve showing detection capabilities of the AE which was pretrained using a WGAN after completion of the training.}
	\label{fig:det}
\end{figure}

It must be noted that the described method was trained on the LivDet2021 Dermalog Sensor dataset where no official test dataset is available yet. The detection capabilities are therefore reported on the holdout validation data set which was also used to track the training process. As those validation data were only used for tracking the reconstruction losses throughout training but not for training itself, nor for defining the classification threshold, the results are comparable to a test dataset. Experimenting with the proposed method on other datasets will be open for future work. The capabilities are reported based on the attack presentation classification error rate (APCER), bona fide presentation classification error rate (BPCER) as well as the average classification error rate (ACR) as defined by ISO/IEC 30107-3 \cite{iso}.
The detection error trade-off (DET) curve shown in Fig.~\ref{fig:det} shows the detection capabilities of the presented approach. As shown in the plot, it seems that the BPCER score was negatively influenced by a few bona fide samples which could not be detected correctly in any case. By setting the classification threshold as described in section \ref{sec:classification}, a ACR of 16.79\%, BPCER of 22.66\% and APCER of 10.9\% could be archived. It must be noted that due to the lack of a missing test dataset there was no hyperparameter optimization involved for this approach as this would have potentially let the model to overfit to the validation data and thus nullify the reported results. Nevertheless it can be expected that the detection capabilities could be further improved by searching for better hyperparameters.

\subsection{Generation of synthetic Fingerprint patches}\label{sec:generation}
\begin{figure}[t]
	\centering
	\includegraphics[scale=0.5]{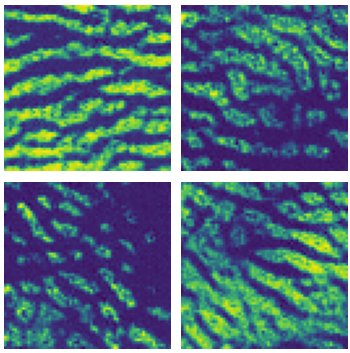}
	\caption{Four synthetic fingerprint patches generated by the Generator of the WGAN based on four random 100 dimensional normally distributed input vectors.}
	\label{fig:gan_created}
\end{figure}

The main application of the GANs in this work was to pretrain the AE as well as to test the detection capabilities of the discriminator as proposed in \cite{gan_fingerprint}. As a side product of this main purpose, the trained WGANs generator can be used to generate a wide variety of realistic synthetic fingerprint patches. During training, the generator learned to generate fingerprint patches based on a 100 dimensional vector randomly drawn from a normal standard distribution. To generate a new random synthetic fingerprint patch, one just has to draw a 100 dimensional standard normal distributed vector and feed it to the generator. An example of four generated images based on four 100 dimensional normal distributed vectors is shown in Fig.~\ref{fig:gan_created}. In case of the WGAN, the generator also supports so called smooth transitions in the latent encoding. Meaning, that two similar 100 dimensional input vectors will generate two similar looking synthetic fingerprints. The functionality and stability of the WGAN was determined by visually checking the quality and variety generated fingerprint patches, as well as to ensure the existence of smooth transitions by manually checking the impact of small changes in the latent encoding on the generated fingerprint patch. The DCGAN without the usage of Wasserstein loss had the issue of mode collapse and was thus not able to generate a wide variety of fingerprint patches.

\subsection{Classification using GAN only}
Most GAN projects discard the discriminator after training to solely make use of the generator. In contrast to this, a concept of using the GANs discriminator for fingerprint PAD was proposed in \cite{gan_fingerprint} by applying the following main ideas:
(i) The generator attempts to create synthetic fingerprints which look like bona fide (ii) The discriminator is trained to discriminate between bona fide and synthetic fingerprints generated by the generator (iii) By the adversarial training process, the generator learns to generate more and more realistic looking synthetic fingerprints to fool the discriminator. The discriminator gets better in discriminating between synthetic and bona fide fingerprints. This could push the discriminator to learn bona fide representations which can be used for discrimination. The generator in this case can be seen as a creative attacker, trying to fool an increasingly improving discriminator \cite{gan_fingerprint} .

\begin{figure}[t]
	\centering
	\includegraphics[scale=0.63]{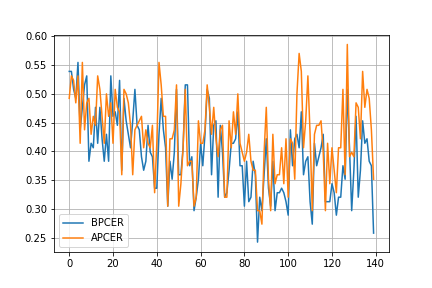}
	\caption{Detection capabilities of the WGAN while it was trained for 140 epochs.}
	\label{fig:gan_deteciton}
\end{figure}

This concept was implemented and tested using the WGAN and DCGAN by retraining them from scratch for 140 epochs by using the same data for training and validation as described in section \ref{sec:dataset}. During training, the detection capabilities were recorded after every epoch to get a better understanding of the progress of the detection capabilities during the adversarial training of the GANs. Classification was done by dividing the input fingerprint in non overlapping patches. All patches were classified individually by the GANs discriminator. A final classification was made by calculating the average of those non overlapping patch classifications. As shown in Fig.~\ref{fig:gan_deteciton}, the detection capabilities fluctuate heavily during training. One hypothesis is, that this fluctuations could be caused by mode collapse. However, mode collapse could be excluded in case of the WGAN by ensuring the GANs stability and functionality after training. This was done by visually checking the variability in generated fingerprints by the generator and by testing the existence of smooth transitions in the latent encoding as described in section \ref{sec:generation}. As the fluctuations during training exist for the DCGAN as well as for the WGAN, the cause of mode collapse can be excluded.

Even after heavy experimenting, the concept of using solely the GANs discriminator for classification as reported in \cite{gan_fingerprint} could not be reproduced. Even not with a stable and healthy WGAN, which learned to create a wide variety of synthetic fingerprint patches and supports smooth transitions in the latent encoding.

\subsection{Convolutional autoencoder from scratch}

Multiple convolutional autoencoder architectures without the use of transfer learning were tested in this work. It must be noted, that training deep convolutional autoencoders showed to be challenging as they are sensitive to small changes in hyperparameters. Additionally, reliable architectural guidelines are only available for GANs \cite{dcgan} but not for deep convolutional autoencoders. By experimenting heavily with different architectures and hyperparameters the following settings turned out to work comparably well in this work.

\begin{enumerate}
	\item Usage of Batch Normalization \cite{bn} after each convolutional layer accelerated convergence by a factor of 10.
	\item Usage of Sigmoid activation function in the last layer of the decoder was necessary for the AE to work at all.	
	\item The usage of strides instead of pooling in both the encoder and decoder which enables to AE to learn the downsampling operation itself \cite{fullyConv}.
	\item The usage of a fully convolutional instead of a fully connected bottleneck improved performance. 
\end{enumerate}

Even after heavy experimenting and implementation of the listed points above, the approach of training an AE from scratch with the little data available could only archive a ACR of 23\% using the same classification strategies as they were reported in section \ref{sec:detection}. Additionally a variational AE \cite{variationalAE} was implemented but did not show promising results during first tests. Further experiments using different architectures especially using variational AEs are recommended and open for future work.

\section{Conclusion}
The concept of pretraining deep convolutional autoencoders using GANs has been presented in this work and showed promising results. Although, this work does not meet the requirements for field deployments with an ACR of 16.79\%, the general applicability of the presented method could be shown by training an one-class classifier for fingerprint presentation attack detection using not more than 1122 bona fide samples. It must be noted, that during training not a single PA sample is required, which differentiates the proposed method from most of the related work. A further differentiation from other one-class classifiers for fingerprint presentation attack detection is again the fact, that the model was trained on only a few yet complex and high dimensional fingerprint image training samples. The WGAN used for pretraining the autoencoder showed to be able to create a wide variety of synthetic fingerprint patches which can be used in future work.

\emph{TensorFlow} implementations of a data preprocessing pipeline, a DCGAN, a WGAN, multiple deep convolutional autoencoders, variational autoencoders as well as the application of transfer learning have been done in the context of this work and are available for future research. For further validation and evaluation of the presented method, it is highly recommended to test it on different datasets as well as to finetune the presented architectures to further improve detection capabilities.

\bibliographystyle{abbrv}
\bibliography{conference_101719}

\end{document}